\documentclass[11pt,a4paper]{article}
\usepackage[hyperref]{acl2019}
\usepackage{times}
\usepackage{latexsym}
\usepackage{url}
\usepackage{paralist}
\usepackage{graphicx}

\aclfinalcopy 

\title{Comparison of Classical Machine Learning Approaches on Bangla Textual Emotion Analysis}

\author{Md. Ataur Rahman \\
  Language Science and Technology \\
  University of Saarland \\
  Saarbr{\"u}cken, Germany \\
  \texttt{arahman@coli.uni-saarland.de} \\\And
  Md. Hanif Seddiqui \\
  Computer Science and Engineering \\
  University of Chittagong \\
  Chittagong, Bangladesh \\
  \texttt{hanif@cu.ac.bd} \\}

\date{}

\begin{document}
\maketitle
\begin{abstract}
  Detecting emotions from text is an extension of simple sentiment polarity detection. Instead of considering only positive or negative sentiments, emotions are conveyed using more tangible manner; thus, they can be expressed as many shades of gray. This paper manifests the results of our experimentation for fine-grained emotion analysis on Bangla text. We gathered and annotated a text corpus consisting of user comments from several Facebook groups regarding socio-economic and political issues, and we made efforts to extract the basic emotions \textit{(sadness, happiness, disgust, surprise, fear, anger)} conveyed through these comments. Finally, we compared the results of the five most popular classical machine learning techniques namely Na\"{\i}ve Bayes, Decision Tree, k-Nearest Neighbor (k-NN), Support Vector Machine (SVM) and K-Means Clustering with several combinations of features. Our best model (SVM with a non-linear radial-basis function (RBF) kernel) achieved an overall average accuracy score of \textit{52.98}\% and an F1 score (macro) of \textit{0.3324}.
\end{abstract}

\section{Introduction}

Sentiment analysis or opinion mining is the task of automatically analyzing text documents using computational methods to obtain the opinions of the authors about specific entities, such as people, companies, events or products. At present, the web has become an excellent source of opinions about entities, particularly with the increased popularity of social media. People are expressing their opinions through reviews, forum discussions, blogs, tweets, comments and posts. Individuals and organizations are increasingly using these opinions for decision-making purposes. 



Thus we took an initiative that aims at developing and annotating a text corpus in Bangla for doing fine-grained emotion analysis. The term \textit{Emotion Analysis} is used because instead of dividing the corpus based on only positive and negative sentiments, we tried to consider more fine-grained emotion labels such as sadness, happiness, disgust, surprise, fear and anger - which are, according to Paul Ekman \shortcite{ekman1999basic}, the six basic emotion categories. Next, we tried to implement five different classical machine learning algorithms, namely the Na\"{\i}ve Bayes, Decision Tree, k-Nearest Neighbours, Support Vector Machine and K-Means clustering on our corpus. Thus the contributions of this paper can briefly be seen in three major folds:

\begin{compactenum}
  \item We will present a manually annotated Bangla emotion corpus, which incorporates the diversity of fine-grained emotion expressions in social-media text.

  \item We will employ classical machine-learning approaches that typically perform well in classifying the six aforementioned emotion types.
  
  \item We will compare the machine-learning classiﬁers’ performance with a baseline to identify the best-performing model for fine-grained emotion classification.
\end{compactenum}

Using our own carefully curated gold standard corpus, we will report our preliminary efforts to train and evaluate machine learning models for emotion classification in Bangla text. Our experimental results show that a non-linear SVM achieved the best performance with an accuracy score of \textit{0.5298}, and an F-score of \textit{0.3324} (macro) and \textit{0.476} (micro) among all the tested classifiers.

\section{Related Works}

To our knowledge, the reliable literature on fine-grained emotion tagging for Bangla is very limited. For example, a reliable research work was that of Das and Bandyopadhyay \shortcite{das2010labeling}. In their work, they annotated a random collection of 123 blog posts consisting a total of 12,149 sentences. The task was mainly focused on observing the performance of different machine learning classifiers. On a small subset of 200 test sentences, the Conditional Random Field (CRF) achieved an average accuracy of \textit{0.587} whereas the SVM scored \textit{0.704}.

In a different paper \cite{das2010developing}, the authors described the preparation of the Bengali WordNet Affect containing six types of emotion words. They employed an automatic method of sense disambiguation. The Bengali WordNet Affect could be useful for emotion-related language processing tasks in Bengali.

In his paper, Das \shortcite{dasanalysis} delineates the identification of the emotional contents at a document-level along with their associated holders and topics. Additionally, he manually annotated a small corpus. By applying sense-based affect-estimation techniques, he gained a micro F-score of \textit{0.66} and \textit{0.619} in terms of `emotion holder' and `emotion topic' identification.

On a case study for Bengali \cite{das2012emotion}, the authors considered 1,100 sentences on eight different topics. They prepared a knowledge base for emoticons and also employed a morphological analyzer to identify the lexical keywords from the Bengali WordNet Affect lists. They claimed an overall precision, recall and F1-Score (micro) of \textit{0.594}, \textit{0.65} and \textit{0.653} respectively.


Jasy and Howard \shortcite{liew2016exploring} investigated certain prevalent machine learning techniques on coarse-grained emotions for English. They used the grounded-theory method to construct a corpus of 5,553 tweets, manually annotated with 28 emotion categories. They showed that SVM and BayesNet outperformed all the classifiers. The BayesNet correctly predicted roughly 60\% of the instances, whereas the SVM was correct on 50\% of the cases.

\section{Data Set} \label{dataset}

We used two different datasets in our experiment. The first was the \textit{Part-of-Speech (POS) Tagset: Bengali} \cite{dandapat2009complex} for POS-tagging. The \textit{dataset}\footnote{\url{https://github.com/abhishekgupta92/bangla_pos_tagger/tree/master/data}} that we were able to obtain contained approximately 3K sentences and 42K words in it's original form, and it contained a broad category of 32 \textit{tagsets}\footnote{\url{http://www.ldcil.org/Download/Tagset/LDCIL/2Bengali.pdf}}.

For the task of the Bangla emotion classification, we annotated 6,314 comments from three different Facebook groups. These comments were mostly reactions to ongoing socio-political issues and concerned the success and failure of the Bangladesh government.

\subsection{Distribution of the Emotion Dataset}

For the purpose of our experiment, we took a balanced set from the aforementioned data and divided it into a training and a test set of an equal ratio. We considered a proportion of 5:1 for training and evaluation purposes. Table \ref{tab2} summarizes this distribution.

\begin{table}[ht]
\centering
\begin{tabular}{|c|c|c|}
\hline
\textbf{\textit{Labels}}& \textbf{\textit{Training Set}}& \textbf{\textit{Testing Set}} \\
\hline
sad & 1000 & 200 \\
\hline
happy & 1500 & 300 \\
\hline
disgust & 500 & 100 \\
\hline
surprise & 400 & 80 \\
\hline
fear & 300 & 60 \\
\hline
angry & 1000 & 200 \\
\hline
\textbf{\textit{Total}} & 4700 & 940 \\
\hline
\end{tabular}
\caption{Distribution of emotion classes in the dataset.}
\label{tab2}
\vspace{-4mm}
\end{table}

\section{Experimental Settings and Methodology}

In this section, we will describe the preprocessing and feature-selection techniques, including the POS tagging approach that we have considered for the emotion recognition models. Finally, the baseline setting for evaluation and further model optimization will be introduced.

\subsection{Preprocessing and Features} \label{preprocess-feature}

Apart from cleaning the data, we also used certain simple text preprocessing techniques. We tokenized the words using a specialized tokenizer for Bangla from spaCy \cite{spacy2}. Moreover, we experimented by filtering out stop words.

We have explored two types of feature vectors namely the \textit{count} vectorizer and a \textit{tf-idf} vectorizer with a combination of \textit{n-grams} (ranging from unigrams to trigrams) from scikit-learn \cite{scikit-learn}. Furthermore, we investigated the effect of POS tagging for feature reduction on our best model.



\subsection{POS Tagging}

For the purpose of POS tagging, we implemented a H\textit{idden Markov Model} (HMM) based tagger. The original POS tagger is capable of tagging 32 tags with an accuracy of \textit{75}\% over the Bangla dataset (Section \ref{dataset}). By looking upon several combinations, we only considered five tags (\textit{`JJ', `CX', `VM', `NP' and `AMN'}) that were the most significant to emotion-related words.

\subsection{Baseline Classifier} \label{baseline-method}

As baseline measure, we used a k-NN classifier with word \textit{unigrams} plus \textit{count} as features. The value of k-nearest neighbours was set to 15 (\textit{k=15}). For evaluation, we will compare the results of this baseline model with the optimized model for each of the classifiers in Section \ref{results}.

  
  
  
  
  

\section{Results and Evaluation} \label{results}

For the evaluation, we will first delineate the results of our baseline classifier (k-NN). Then, we will attempt to find the best model based on the results obtained in Section \ref{baseline-results} through \ref{svm-results}.

\subsection{Baseline Performance} \label{baseline-results}

Table \ref{tab3} lists in detail the results of the baseline model, whereas Table \ref{tab4} summarizes the overall accuracy and the F1(macro) score.

\begin{table}[ht!]
\centering
\begin{tabular}{|c|c|c|c|}
\hline
\textbf{\textit{Labels}}& \textbf{\textit{Precision}}& \textbf{\textit{Recall}}& \textbf{\textit{F1(micro)}} \\
\hline
angry   &   0.125  &   0.020  &   0.034 \\
\hline
disgust   &   1.000   &  0.010   &  0.020 \\
\hline
fear   &   0.000   &  0.000  &   0.000 \\
\hline
happy   &   \textbf{0.342}  &   \textbf{0.977}   &  \textbf{0.507} \\
\hline
sad   &   0.421  &   0.040   &  0.073 \\
\hline
surprise   &   0.125  &   0.037  &   0.058 \\
\hline
\textbf{\textit{Average}} & 0.342  &   0.329  &   0.192 \\
\hline
\end{tabular}
\caption{Results of K-NN Classifier as the Baseline Model with \textit{K=15}.}
\label{tab3}
\vspace{-4mm}
\end{table}

\begin{table}[ht!]
\centering
\begin{tabular}{|c|c|}
\hline
\textbf{\textit{Evaluation Matrix}}& \textbf{\textit{Score}} \\
\hline
\textbf{Accuracy}   &   0.329  \\
\hline
\textbf{F1(macro)}   &   0.115  \\
\hline
\end{tabular}
\caption{Average score of the Baseline Model.}
\label{tab4}
\vspace{-4mm}
\end{table}

\begin{figure}[ht!]
    \centering
    \includegraphics[width=0.45\textwidth]{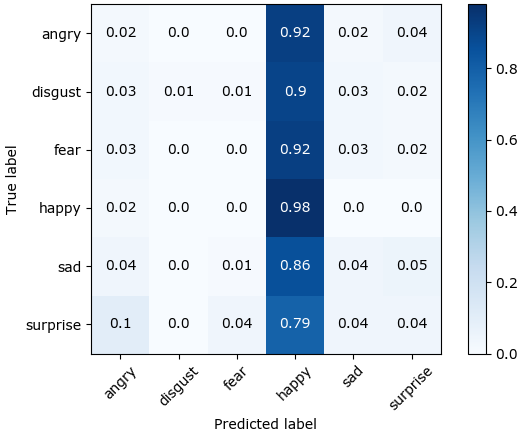}
    \caption{Confusion Matrix (Accuracy) for Baseline.}
    \label{fig1}
    \vspace{-4mm}
\end{figure}

From Table \ref{tab3} and Fig. \ref{fig1}, it may be observed that the baseline classifier predicts almost every class as being \textit{`happy'}. This could be the result of the classifier being biased towards this particular label because the maximum number in the training example was supplied for the category \textit{`happy'}.

\subsection{K-Nearest Neighbours}

\begin{figure}[ht!]
\centerline{\includegraphics[width=0.45\textwidth]{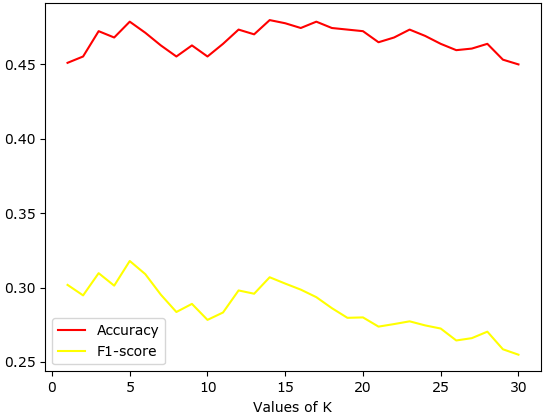}}
\caption{Plot of accuracy and F1-score(macro) for different values of k.}
\label{fig2}
\vspace{-4mm}
\end{figure}

Although the baseline k-NN model performed quite poorly, we attempted to tune the parameters for the K-NN classifier to identify the best k-value for our data. Table \ref{tab5} and Fig. \ref{fig2} presents the results of the k-NN classifier for various k-values. It should be noted that here, we only considered the \textit{tf-idf} feature because it yielded better results than the \textit{count} feature. Considering the data and the plot (Fig \ref{fig2}), it is obvious that the classifier produces the best outputs for the k value of \textit{5}.

\begin{table}[ht!]
\centering
\begin{tabular}{|c|c|c|}
\hline
\textbf{\textit{K-Values}}& \textbf{\textit{Accuracy}}& \textbf{\textit{F1(macro)}} \\
\hline
2  &  0.455  &   0.295  \\
\hline
3  &  0.472  &   0.31  \\
\hline
4  &  0.468  &   0.301  \\
\hline
\textbf{5}  &  \textbf{0.479}  &   \textbf{0.318}  \\
\hline
6  &  0.471  &   0.309  \\
\hline
7  &  0.463  &   0.295  \\
\hline
\end{tabular}
\caption{Results of K-NN Classifier with Tf-Idf Feature for Different k-values.}
\label{tab5}
\vspace{-4mm}
\end{table}

We selected the value of \textit{k=5} as our default parameter to be further examined with different preprocessing and feature combinations (Table \ref{tab6}). The results indicate that the best k-NN model uses the tf-idf unigram as a feature (\textit{accuracy = 0.479}, \textit{F1-macro = 0.318}). 

\begin{table}[ht!]
\centering
\begin{tabular}{|p{0.42\linewidth}|p{0.18\linewidth}|p{0.22\linewidth}|}
\hline
\textbf{\textit{Feature}}& \textbf{\textit{Accuracy}}& \textbf{\textit{F1(macro)}} \\
\hline
unigram + count  &  0.359  &   0.172  \\
\hline
\textbf{unigram + tf-idf}  &  \textbf{0.479}  &   \textbf{0.318}  \\
\hline
stopword + tf-idf  &  0.332  &   0.133  \\
\hline
stopword + count  &  0.342  &   0.146  \\
\hline
stopword + tf-idf + n-gram(1,3)  &  0.316  &   0.091  \\
\hline
stopword + count + n-gram(1,3)  &  0.330  &   0.114  \\
\hline
\end{tabular}
\caption{Results of K-NN Classifier for Different Feature Combinations (K=5).}
\label{tab6}
\vspace{-4mm}
\end{table}



\subsection{Na\"{\i}ve Bayes}

For our second classification algorithm, we used the `\textit{Multinomial Naive Bayes}' (MNB) classifier. Unlike certain other classifiers, the MNB did not require setting and tuning the parameters. Thus, we directly experimented with different feature/preprocessing techniques (Table \ref{tab8}).

\begin{table}[ht!]
\centering
\begin{tabular}{|p{0.42\linewidth}|p{0.18\linewidth}|p{0.22\linewidth}|}
\hline
\textbf{\textit{Feature}}& \textbf{\textit{Accuracy}}& \textbf{\textit{F1(macro)}} \\
\hline
unigram + tf-idf  &  0.491  &   0.266  \\
\hline
\textbf{unigram + count}  &  \textbf{0.525}  &   \textbf{0.295}  \\
\hline
stopword + tf-idf  &  0.472  &   0.250  \\
\hline
stopword + Count  &  0.506  &   0.284  \\
\hline
n-gram(1,3) + tf-idf  &  0.444  &   0.227  \\
\hline
n-gram(1,3) + count  &  0.516  &   0.287  \\
\hline
stopword + tf-idf + n-gram(1,3)  &  0.434  &   0.219  \\
\hline
stopword + count + n-gram(1,3)  &  0.515  &   0.292  \\
\hline
\end{tabular}
\caption{Results of Multinomial Na\"{\i}ve Bayes Classifier for Different Feature Combinations.}
\label{tab8}
\vspace{-4mm}
\end{table}

Based on the above results, the best MNB model was achieved by combining the count with the unigram feature; an accuracy score of \textit{0.525} and an F1-macro of \textit{0.295} were obtained during the test.

\subsection{Decision Tree}

The DT constructs a regression or classification model by following a tree structure. Here, optimal parameter settings were found by considering a \textit{minimum samples split} of \textit{2}, \textit{minimum sample leaf size} of \textit{1}. We did not impose any restrictions on the number of features and to the depth of the tree. Table \ref{tab10} lists the results of several combinations of features and preprocessing schemes.

\begin{table}[ht!]
\centering
\begin{tabular}{|p{0.42\linewidth}|p{0.18\linewidth}|p{0.22\linewidth}|}
\hline
\textbf{\textit{Feature}}& \textbf{\textit{Accuracy}}& \textbf{\textit{F1(macro)}} \\
\hline
\textbf{unigram + tf-idf}  &  \textbf{0.442}  &   \textbf{0.301}  \\
\hline
unigram + count  &  0.432  &   0.287  \\
\hline
stopword + tf-idf  &  0.416  &   0.283  \\
\hline
stopword + count  &  0.430  &   0.292  \\
\hline
stopword + tf-idf + n-gram(1,3)  &  0.394  &   0.247  \\
\hline
stopword + count + n-gram(1,3)  &  0.421  &   0.277  \\
\hline
\end{tabular}
\caption{Results of Decision Tree Classifier for Different Feature Combinations.}
\label{tab10}
\vspace{-4mm}
\end{table}

According to the aforementioned results, the best DT model with an accuracy of \textit{0.442} and an F1(macro) of \textit{0.301} was obtained from the unigram and tf-idf combination.



\subsection{K-Means Clustering}

The only unsupervised machine learning approach that we used in our experiment was the \textit{K-Means Clustering}. We selected a \textit{cluster size} of \textit{N=6}, as we have six different emotion categories. We investigated different initialization ranging from 1 to 15. Here, the initialization (\textit{n\_init}) is the number of times the k-means algorithm executes with different centroid seeds. The final results would be the best output of \textit{n} consecutive runs. To evaluate the clustering, we used two measures: \textit{Adjusted Rand-Index} and \textit{V-measure} (similar to F-measure).


\begin{table}[ht!]
\centering
\begin{tabular}{|p{0.42\linewidth}|p{0.19\linewidth}|p{0.23\linewidth}|}
\hline
\textbf{\textit{Feature}}& \textbf{\textit{Adjusted Rand}}& \textbf{\textit{V-measure}} \\
\hline
unigram + tf-idf  &  0.008  &   0.042  \\
\hline
unigram + count  &  0.009  &   0.009  \\
\hline
\textbf{n-gram (1,3) + tf-idf}  & \textbf{0.059}  &   \textbf{0.049}  \\
\hline
n-gram (1,3) + count  &  0.009  &   0.011  \\
\hline
\end{tabular}
\caption{Results of K-Means Clustering Algorithm for Different Feature Combinations}
\label{tab14}
\vspace{-4mm}
\end{table}

Table \ref{tab14} lists the best evaluation scores for the k-means clustering model (for \textit{n\_init} of 1 to 15). From the results, we can see that the highest score of \textit{0.049} in terms of the V-measure and an Adjusted Rand-Index score of \textit{0.059} was achieved using a combination of \textit{ngram(1,3)} and \textit{tf-idf} feature.

\subsection{Support Vector Machine} \label{svm-results}

To find the best SVM-model, we used both linear and non-linear SVM-kernel. In both cases, the most important words (tf-idf) were used as features because it leads to the highest performance. We explored different values for \textit{Gamma} and \textit{C-parameters} and found that the non-linear kernel performed slightly better. Results and settings for the SVM model will be discussed in Section \ref{best-model}.

\subsection{Best Model} \label{best-model}

Among all models, the best model was the `\textit{SVM with a Non-Linear RBF-Kernel}'. Next, we continued experimenting with different preprocessing and feature combinations using this SVM model. Table \ref{tab7} summarizes an overview of different combinations of feature and preprocessing techniques on the non-linear SVM model with an RBF-kernel. The most optimal parameter settings in this experiment were a C-parameter value of \textit{2.0} and a Gamma-value of \textit{0.6}. The best combination of features was the most important (\textit{tf-idf}) word \textit{unigrams}. Therefore, the highest \textit{accuracy} score achieved from the model was \textit{0.5298} (i.e., an improvement of \textit{20.08}\% from the baseline model) and an \textit{F1(macro)} of \textit{0.3324} (i.e., an improvement of 0.2174 from baseline model).

\begin{table}[ht!]
\centering
\begin{tabular}{|p{0.34\linewidth}|p{0.15\linewidth}|p{0.18\linewidth}|p{0.15\linewidth}|}
\hline
\textbf{\textit{Feature}}& \textbf{\textit{Gamma}}& \textbf{\textit{Accuracy}}& \textbf{\textit{F1 (macro)}} \\
\hline
POS + unigram + tf-idf  &   0.6 &   0.399  &   0.226  \\
\hline
\textbf{unigram + tf-idf}  &   \textbf{0.6} &   \textbf{0.5298}  &   \textbf{0.3324}  \\
\hline
unigram + stopword + tf-idf  &  0.3  &  0.517  &   0.312  \\
\hline
n-gram(1,3) + tf-idf  &  0.8  &   0.516  &   0.307  \\
\hline
n-gram(1,3) + tf-idf + POS  &  0.8  &   0.399  &   0.224  \\
\hline
n-gram(1,3) + stopword + tf-idf  &  0.4  &  0.525  &   0.313  \\
\hline
n-gram(1,3) + stopword + tf-idf + POS &  0.4  &  0.374  &   0.186  \\
\hline
Feature Union   &  0.6  &  0.322  &   0.087  \\
\hline
\end{tabular}
\caption{Results of Best Model for Different Features.}
\label{tab7}
\vspace{-2mm}
\end{table}

\begin{figure}[ht!]
    \centering
    \includegraphics[width=0.45\textwidth]{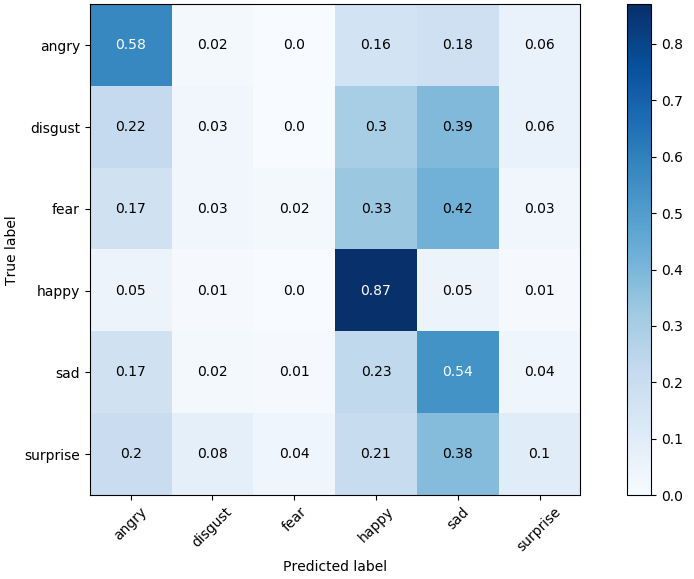}
    \caption{Confusion Matrix (Accuracy) for Best SVM model with unigram and tf-idf as features}
    \label{fig5}
    \vspace{-4mm}
\end{figure}

Based on this best combination of features listed in Table \ref{tab7}, the detailed results are delineated in Table \ref{tab9}. Again, we may obtain a clear insight into the model's prediction or misclassification from the confusion matrix illustrated in Fig. \ref{fig5}.

\begin{table}[ht!]
\centering
\begin{tabular}{|c|c|c|c|}
\hline
\textbf{\textit{Labels}}& \textbf{\textit{Precision}}& \textbf{\textit{Recall}}& \textbf{\textit{F1(micro)}} \\
\hline
angry   &   0.547   &  0.585  &   0.565 \\
\hline
disgust   &   0.136   &  0.030   &  0.049 \\
\hline
fear   &   0.143  &   0.017  &   0.030 \\
\hline
happy   &   0.645  &   0.873   &  0.742 \\
\hline
sad   &   0.425  &   0.535   &  0.473 \\
\hline
surprise   &   0.205  &   0.100   &  0.134 \\
\hline
\textbf{\textit{Average}} & 0.454  &   0.530   &  0.476 \\
\hline
\end{tabular}
\caption{Detailed Results of Best Model with Unigram and Tf-idf as Features}
\label{tab9}
\vspace{-4mm}
\end{table}

\section{Discussion and Conclusions}

The linguistic motivation behind this project was inspired by the growing field of computational research in natural languages, particularly in the Bangla language processing because Bangla is one of the most widely spoken languages; it ranked 7th in the world, with staggering number of 268 million native speakers. However, the computational motivation was to compare the contribution of different features on the performance of a classifier in doing fine-grained Bangla emotion analysis. The findings of this study imply that the SVM model that best predicted the aforementioned emotions in Bangla text composed of social-media comments, was a model that used a non-linear RBF-kernel, which yielded an \textit{accuracy} of \textit{52.98}\% and an \textit{F1-score} of \textit{0.3324} (macro) and \textit{0.476} (micro). These scores showed a significant improvement over the \textit{Baseline} model, with nearly a \textit{20.08}\% increase in \textit{accuracy}. Additionally, both the \textit{F1} macro and micro scores increased by \textit{0.217} and \textit{0.284}, respectively.

\bibliography{acl2019}
\bibliographystyle{acl_natbib}

\end{document}